\documentclass[sigplan,screen]{acmart}
\AtBeginDocument{%
  \providecommand\BibTeX{{%
    \normalfont B\kern-0.5em{\scshape i\kern-0.25em b}\kern-0.8em\TeX}}}
\setcopyright{none}
\begin{document}
\settopmatter{printacmref=false}
\renewcommand\footnotetextcopyrightpermission[1]{}
\pagestyle{plain}
\title{The Science of Detecting LLM-Generated Texts}


\author{Ruixiang Tang, Yu-Neng Chuang, Xia Hu}
\affiliation{%
  \institution{Department of Computer Science, Rice University}
  \streetaddress{6100 Main St}
  \city{Houston}
  \country{USA}}
\email{{rt39, ynchuang, xia.hu }@rice.edu}


\begin{abstract}
The emergence of large language models (LLMs) has resulted in the production of LLM-generated texts that is highly sophisticated and almost indistinguishable from texts written by humans. However, this has also sparked concerns about the potential misuse of such texts, such as spreading misinformation and causing disruptions in the education system. Although many detection approaches have been proposed, a comprehensive understanding of the achievements and challenges is still lacking. This survey aims to provide an overview of existing LLM-generated text detection techniques and enhance the control and regulation of language generation models. Furthermore, we emphasize crucial considerations for future research, including the development of comprehensive evaluation metrics and the threat posed by open-source LLMs, to drive progress in the area of LLM-generated text detection.
\end{abstract}
\maketitle



\begin{figure*}[t]
\centering
\includegraphics[width=0.9\textwidth]{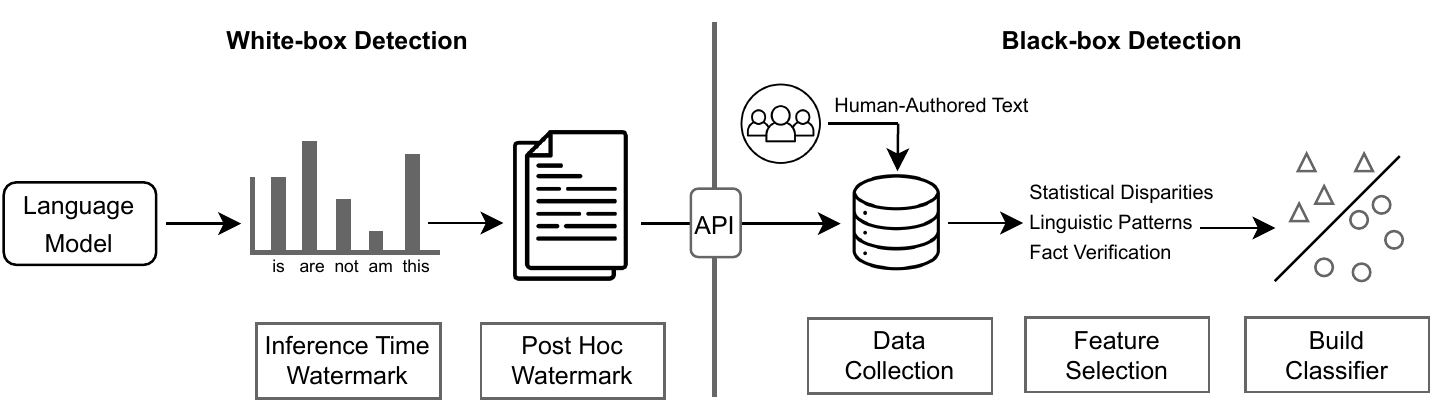}
 \caption{An overview of the LLM-generated text detection.}
 \label{fig: pipeline}
\end{figure*}

\section{Introduction}
Recent advancements in natural language generation (NLG) technology have significantly improved the diversity, control, and quality of LLM-generated texts. A notable example is OpenAI's ChatGPT, which demonstrates exceptional performance in tasks such as  answering questions, composing emails, essays, and codes. However, this newfound capability to produce human-like texts at high efficiency also raises concerns about detecting and preventing misuse of LLMs in tasks such as phishing, disinformation, and academic dishonesty. For instance, many schools banned ChatGPT due to concerns over cheating in assignments \cite{chalkbeat2023ban}, and media outlets have raised the alarm over fake news generated by LLMs \cite{floridi2020gpt}. These concerns about the misuse of LLMs have hindered the NLG application in important domains such as media and education. 

The ability to accurately detect LLM-generated texts is critical for realizing the full potential of NLG while minimizing serious consequences. From the perspective of end-users, LLM-generated text detection could increase trust in NLG systems and encourage adoption. For machine learning system developers and researchers, the detector can aid in tracing generated texts and preventing unauthorized use. Given its significance, there has been a growing interest in academia and industry to pursue research on LLM-generated text detection and to deepen our understanding of its underlying mechanisms.

While there is a rising discussion on whether LLM-generat-ed texts could be properly detected and how this can be done, we provide a comprehensive technical introduction of existing detection methods which can be roughly grouped into two categories: black-box detection and white-box detection. Black-box detection methods are limited to API-level access to LLMs. They rely on collecting text samples from human and machine sources, respectively, to train a classification model that can be used to discriminate between LLM- and human-generated texts. Black-box detectors work well because current LLM-generated texts often show linguistic or statistical patterns. However, as LLMs evolve and improve, black-box methods are becoming less effective. An alternative is white-box detection, in this scenario, the detector has full access to the LLMs and can control the model's generation behavior for traceability purposes. In practice, black-box detectors are commonly constructed by external entities, whereas white-box detection is generally carried out by LLM developers.
\balance 

This article is to discuss the timely topic from a data mining and natural language processing perspective. Specifically, we first outline the  black-box detection methods in terms of a data analytic life cycle, including data collection, feature selection, and classification model design. We then delve into more recent advancements in white-box detection methods, such as post-hoc watermarks and inference time watermarks. Finally, we present the limitations and concerns of current detection studies and suggest potential future research avenues. We aim to unleash the potential of powerful LLMs by providing fundamental concepts, algorithms, and case studies for detecting LLM-generated texts. 

\section{Prevalence and Impact} Recent advancements in LLMs, such as OpenAI's ChatGPT, have emphasized the potential impacts of this technology on individuals and society. Demonstrated through its performance on challenging tests, such as the MBA exams at Wharton Business School \cite{chatgpt2023MBA}, the capabilities of ChatGPT suggest its potential to provide professional assistance across various disciplines. Specifically in the healthcare domain, the applications of ChatGPT extend far beyond simple enhancements in efficiency. The ChatGPT not only optimizes documentation procedures, facilitating the generation of medical records, progress reports, and discharge summaries, it aids in the collection and analysis of patient data, facilitating medical professionals in making informed decisions regarding patient care. Recent research has also indicated the potential of LLMs in generating synthetic data for the healthcare field, thereby potentially addressing common privacy issues during data collection. LLMs' influence is also felt in the legal sector, where they are reshaping traditional practices such as contract generation and litigation procedures. The improved efficiency and effectiveness of these models are changing how we do things across a multitude of domains. As a result, when we talk about LLMs, we're not just measuring their technical competency, but also looking at their broader societal and professional implications.

The introduction of LLMs in education has elicited huge concerns. While convenient, their potential to provide quick answers threatens to undermine the development of critical thinking and problem-solving skills, which are essential for academic and life-long success. Further, there's the issue of academic honesty, as students might be tempted to use these tools inappropriately. In response, New York City Public Schools have prohibited the use of ChatGPT \cite{chalkbeat2023ban}. While the impact of LLMs on education is significant, it is imperative to extend this discourse to other domains. For instance, in journalism, the emergence of AI-generated "deepfake" news articles can threaten the credibility of news outlets and misinform the public. In the legal sector, the potential misuse of LLMs could have repercussions on the justice system, from contract generation to litigation processes. In cybersecurity, LLMs could be weaponized to create more convincing phishing emails or social engineering attacks.

In an attempt to mitigate the potential misuse of LLMs, detection systems for LLM-generated text are emerging as a significant countermeasure. These systems offer the capability to differentiate AI-generated content from human-authored text, thereby playing a pivotal role in preserving the integrity of various domains. In the realm of academia, such tools can facilitate the identification of academic misconduct. Within the field of journalism, these systems may assist in separating legitimate news from AI-generated misinformation. Furthermore, in cybersecurity, they have the potential to strengthen spam filters to better identify and flag AI-aided threats. A recent incident at Texas A\&M University underscores the urgent need for effective LLM detection tools \cite{texasam2023instructor}. An instructor suspected students of using ChatGPT to complete their final assignments. Lacking a detection tool, the instructor resorted to pasting the student's responses into ChatGPT, asking the ChatGPT if it had generated the text. This ad-hoc method sparked substantial debate online, illustrating the pressing need for more sophisticated and reliable ways to detect LLM-generated content. While the current detection tools may not be flawless, they nonetheless symbolize a proactive effort to maintain ethical standards in the face of rapid AI advancements. The surge of interest in research focused on LLM-generated text detection testifies to the importance of these tools in mitigating the societal impact of LLMs. As such, we must conduct more extensive discussions on the detection of LLM-generated text. Particularly, we must explore its potential to safeguard the integrity of various domains against the risks posed by LLM misuse.

\section{Black-box Detection}
In the domain of black-box detection, external entities are restricted to API-level access to the LLM, as depicted in Figure \ref{fig: pipeline}. To develop a proficient detector, black-box approaches necessitate gathering text samples originating from both human and machine-generated sources. Subsequently, a classifier is then designed to distinguish between the two categories by identifying and leveraging relevant features. We highlight the three essential components of black-box text detection: data acquisition, feature selection, and the execution of the classification model.

\subsection{Data Acquisition}
The effectiveness of black-box detection models is heavily dependent on the quality and diversity of the acquired data. Recently, a growing body of research has concentrated on amassing responses generated by LLMs and comparing them with human-composed texts spanning a wide range of domains. This section delves into the various strategies for obtaining data from both human and machine sources.

\subsubsection{LLM-generated Data:}

LLMs are designed to estimate the likelihood of subsequent tokens within a sequence, based on the preceding words. Recent advancements in natural language generation have led to the development of LLMs for various domains, including question-answering, news generation, and story creation. Prior to acquiring LM-generated texts, it is essential to delineate target domains and generation models. Typically, a detection model is constructed to recognize text generated from a specific LM across multiple domains. In order to enhance detection generalizability, the minimax strategy suggests that detectors should minimize worst-case performance, which entails enhancing detection capabilities for the most challenging instances where the quality of LM-generated text closely resembles human-authored content \cite{pagnoni-etal-2022-threat}.

Generating high-quality text in a particular domain can be achieved by fine-tuning LMs on task-related data, which substantially improves the quality of the generated texts. For example, Solaiman et al. fine-tuned the GPT-2 model on Amazon product reviews, producing reviews with a style consistent with those found on Amazon \cite{solaiman2019release}. Moreover, LMs are known to produce artifacts such as repetitiveness, which can negatively impact the generalizability of the detection model. To mitigate these artifacts, researchers can provide domain-specific prompts or constraints before generating outputs. For instance, Clark et al. \cite{clark2021all} randomly selected 50 articles from Newspaper3k to use as prompts for the GPT-3 model for news generation and applied filtering constraints on the models with the phrase "Once upon a time" for story creation. The token sampling strategy also significantly influences the generated text quality and style. While deterministic greedy algorithms like beam search \cite{tillmann2003word} generate the most probable sequence, they may restrict creativity and language diversity. Conversely, stochastic algorithms such as nucleus sampling \cite{holtzman2019curious} maintain a degree of randomness while excluding inferior candidates, making them more suitable for a free-form generation. In conclusion, it is crucial for researchers to carefully consider the target domain, generation models, and sampling strategies when collecting LM-generated text to ensure the production of high-quality, diverse, and domain-appropriate content. 

\subsubsection{Human-Authored Data} 
Manual composition by humans serves as a natural method for obtaining authentic, human-authored data. For example, in a study conducted by Dugan et al. \cite{dugan2020roft}, the authors sought to evaluate the quality of natural language generation systems and gauge human perceptions of the generated texts. To achieve this, they employed 200 Amazon Mechanical Turk workers to complete 10 annotations on a website, accompanied by a natural language rationale for their choices. However, manually collecting data through human effort can be both time-consuming and financially impractical for larger datasets. An alternative strategy involves extracting text directly from human-authored sources, such as websites and scholarly articles. For instance, we can readily amass thousands of descriptions of computer science concepts from Wikipedia, penned by knowledgeable human experts \cite{guo2023close}. Moreover, numerous publicly accessible benchmark datasets, like ELI5 \cite{fan2019eli5}, which comprises 270K threads from the Reddit forum "Explain Like I'm Five", already offer human-authored texts in an organized format. Utilizing these readily available sources can considerably decrease the time and expense involved in collecting human-authored texts. Nevertheless, it is essential to address potential sampling biases and ensure topic diversity by including texts from various groups of people, as well as non-native speakers.

\subsubsection{Human Evaluation Findings:} Prior research has offered valuable perspectives on differentiating LLM-generat-ed texts from human-authored texts through human evaluations. Initial observations indicate that LLM-generated texts are less emotional and objective compared to human-authored text, which often uses punctuation and grammar to convey subjective feelings~\cite{guo2023close}. For example, human authors frequently use exclamation marks, question marks, and ellipsis to express their emotions, while LLMs generate answers that are more formal and structured. However, it is crucial to acknowledge that LLM-generated texts may not always be accurate or beneficial, as they can contain fabricated information \cite{shakeel2021fake, guo2023close}. At the sentence level, research has shown that human-authored texts are more coherent than LLM-generated text, which tends to repeat terms within a paragraph~\cite{dugan2022real, pagnoni2022threat}. These observations suggest that LLMs may leave some distinctive signals in their generated text, allowing for the selection of suitable features to distinguish between LLM and human-authored texts.

\subsection{Detection Feature Selection}
How can we discern between LM-generated texts and human-authored texts? This section will discuss possible detection features from multiple angles, including statistical disparities, linguistic patterns, and fact verification.

\begin{figure}[t]
\centering
\includegraphics[width=1.0\columnwidth]{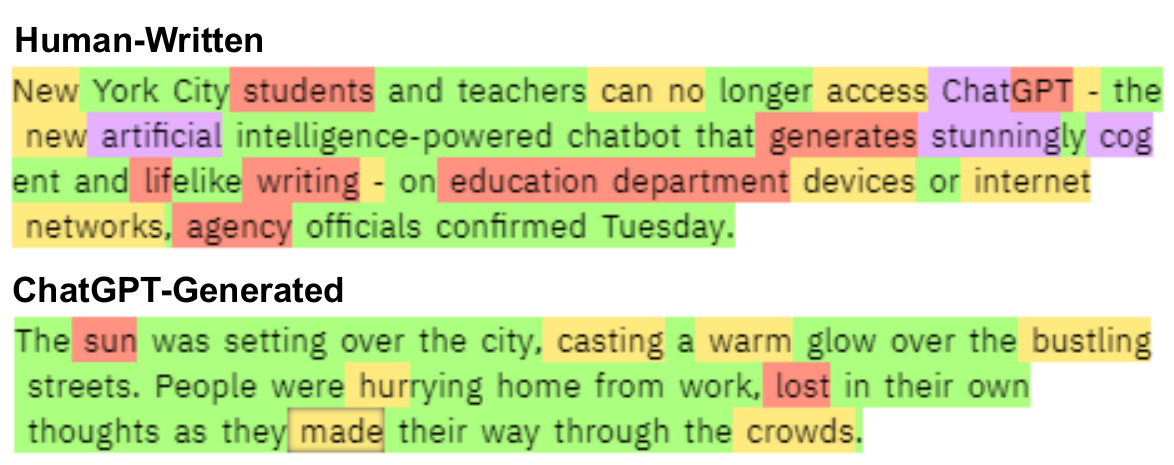}
 \caption{Visualization results of GLTR \cite{gehrmann2019gltr}, the word ranking is obtained from the GPT-2 small model. Words that rank within the top 10 are highlighted in green, top 100 in yellow, top 1,000 in red, and the rest in purple. There is a notable difference between the two texts. The human-authored texts are from Chalkbeat New York \cite{chalkbeat2023ban}.}
 \label{fig: gltr}
\end{figure}

\subsubsection{Statistical Disparities}
\begin{figure}[t]
\centering
\includegraphics[width=1.0\columnwidth]{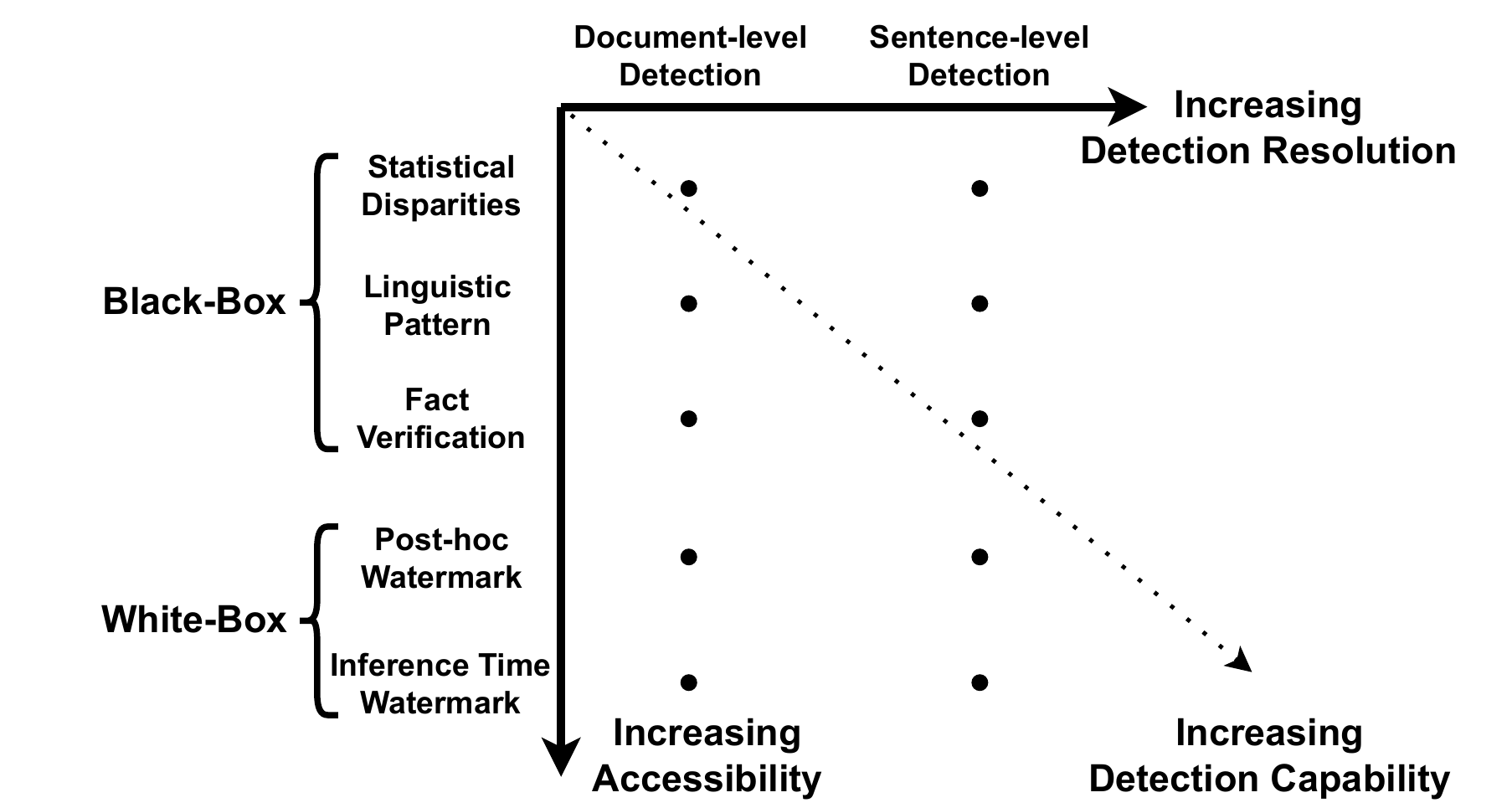}
 \caption{A taxonomy of LLM-generated text detection.}
 \label{fig: overview}
\end{figure}

The detection of statistical disparities between LLM-generated and human-authored texts can be accomplished by employing various statistical metrics. For example, the Zipfian coefficient measures the text's conformity to an exponential curve, which is described by Zipf's law~\cite{piantadosi2014zipf}. A visualization tool, GLTR \cite{gehrmann2019gltr}, has been developed to detect generation artifacts across prevalent sampling methods, as demonstrated in Figure \ref{fig: gltr}. The underlying assumption is that most systems sample from the head of the distribution, thus word ranking information of the language model can be used to distinguish LLM-generated text. Perplexity serves as another widely used metric for LLM-generated text detection. It measures the degree of uncertainty or surprise in predicting the next word in a sequence, based on the preceding words, by calculating the negative average log-likelihood of the texts under the language model \cite{brown1992estimate}. Research indicates that language models tend to concentrate on common patterns in the texts they were trained on, leading to low perplexity scores for LLM-generated text. In contrast, human authors possess the capacity to express themselves in a wide range of styles, making prediction more difficult for language models and resulting in higher perplexity values for human-authored texts. However, it is important to recognize that these statistical disparities are constrained by the necessity for document-level text, which inevitably diminishes the detection resolution, as depicted in Figure \ref{fig: overview}.

\subsubsection{Linguistic Patterns} 
Various contextual properties can be employed to analyze linguistic patterns in human and LLM-generated texts, such as vocabulary features, part-of-speech, dependency parsing, sentiment analysis, and stylistic features. The vocabulary features offer insight into the queried text's word usage patterns by analyzing characteristics such as average word length, vocabulary size, and word density. Previous studies on ChatGPT have shown that human-authored texts tend to have a more diverse vocabulary but shorter length \cite{guo2023close}. Part-of-speech analysis emphasizes the dominance of nouns in ChatGPT texts, implying argumentativeness and objectivity, while the dependency parsing analysis shows that ChatGPT texts use more determiners, conjunctions, and auxiliary relations \cite{guo2023close}. Sentiment analysis, on the other hand, provides a measure of the emotional tone and mood expressed in the text. Unlike humans, large language models tend to be neutral by default and lack emotional expression. Research has shown that ChatGPT expresses significantly less negative emotion and hate speech compared to human-authored texts. Stylistic features or stylometry, including repetitiveness, lack of purpose, and readability, are also known to harbor valuable signals for detecting LLM-generated texts \cite{frohling2021feature}. In addition to analyzing single texts, numerous linguistic patterns can be found in multi-turn conversations \cite{bhatt2021detecting}. These linguistic patterns are a reflection of the training data and strategies of LLMs and serve as valuable features for detecting LLM-generated text. However, it is important to note that LLMs can substantially alter their linguistic patterns in response to prompts. For instance, incorporating a prompt like "Please respond with humor" can change the sentiment and style of the LLM's response, impacting the robustness of linguistic patterns.  

\subsubsection{Fact Verification}
Large Language models often rely on likelihood maximization objectives during training, which can result in the generation of nonsensical or inconsistent text, a phenomenon known as hallucination. This emphasizes the significance of fact-verification as a crucial feature for detection \cite{zhong2020neural}. For instance, OpenAI's ChatGPT has been reported to generate false scientific abstracts and post misleading news opinions. Studies also revealed that popular decoding methods, such as top-k and nucleus sampling, resulted in more diverse and less repetitive generations. However, they also produce texts that are less verifiable ~\cite{massarelli2020decoding}. These findings underscore the potential for using fact verification to detect LLM-generated texts.

Prior research has advanced the development of tools and algorithms for conducting fact verification, which entails retrieving evidence for claims, evaluating consistency and relevance, and detecting inconsistencies in texts. One strategy employs sentence-level evidence, such as extracting facts from Wikipedia, to directly verify facts of a sentence \cite{massarelli2020decoding}.  Another approach involves analyzing document-level evidence via graph structures, which capture the factual structure of the document as an entity graph. This graph is utilized to learn sentence representations with a graph neural network, followed by the composition of sentence representations into a document representation for fact verification \cite{zhong2020neural}. Some studies also use knowledge graphs constructed from truth sources, such as Wikipedia, to conduct fact verification \cite{shakeel2021fake}. These methods evaluate consistency by querying subgraphs and identify non-factual information by iterating through entities and relations. Given that human-authored texts may also contain misinformation, it is vital to supplement the detection results with other features in order to accurately distinguish texts generated by LLMs.

\subsection{Classification Model}
The detection task is typically approached as a binary classification problem, aiming to capture textual features that differentiate between human-authored and LLM-generated texts. This section provides an overview of the primary categories of classification models. 


\subsubsection{Traditional Classification Algorithms}
Traditional classification algorithms utilize various features outlined in Section 3.2 to differentiate between human-authored and LLM-generated texts. Some of the commonly used algorithms are Support Vector Machines, Naive Bayes, and Decision Trees. For instance, Fröhling et al. utilized linear regression, SVM, and random forests models built on statistical and linguistic features to successfully identify texts generated by GPT-2, GPT-3, and Grover models \cite{frohling2021feature}. Similarly, Solaiman et al. achieved solid performance in identifying texts generated by GPT-2 through a combination of TF-IDF unigram and bigram features with a logistic regression model \cite{solaiman2019release}. In addition, studies have also shown that using pre-trained language models to extract semantic textual features, followed by SVM for classification, can outperform the use of statistical features alone \cite{crothers2022adversarial}. One advantage of these algorithms is their interpretability, allowing researchers to analyze the importance of input features and understand why the model classifies texts as LLM-generated or not.



\subsubsection{Deep Learning Approaches}
In addition to employing explicitly extracted features for detection, recent studies have explored leveraging language models, such as RoBERTa \cite{liu2019roberta}, as a foundation. This approach entails fine-tuning these language models on a mixture of human-authored and LLM-generated texts, enabling the implicit capture of textual distinctions. The majority of studies adopt the supervised learning paradigm for training the language model, as demonstrated by Ippolito et al. \cite{ippolito2020automatic}, who fine-tuned the BERT model on a curated dataset of generated-text pairs. Their study revealed that human raters have significantly lower accuracy than automatic discriminators in identifying LLM-generated text. In a low-resource scenario, Rodriguez et al. \cite{rodriguez2022cross} showed that a few hundred labeled in-domain authentic and synthetic texts suffice for robust performance, even without complete information about the LLM text generation pipeline. Despite the strong performance under the supervised learning paradigms, obtaining annotations for detection data can be challenging in real-world applications, leading the supervised paradigms inapplicable in some cases. Recent research~\cite{galle2021unsupervised} addresses this issue by detecting LLM-generated documents using repeated higher-order n-grams, which can be trained under unsupervised learning paradigms without requiring LLM-generated datasets as training data. Besides using the language model as the backbone, recent research finds that contextual structure can be viewed as a graph containing entities mentioned in the texts and the semantically relevant relations, which utilizes a deep graph neural network to capture the structure feature of a document for LLM-generated news detection \cite{zhong2020neural}. While deep learning approaches often yield superior detection outcomes, their black-box nature severely restricts interpretability. Consequently, researchers typically rely on interpretation tools to comprehend the rationale behind the model's decisions.

\section{White-box Detection}

In white-box detection, the detector processes complete access to the target language model, facilitating the integration of concealed watermarks into its outputs to monitor suspicious or unauthorized activities. In this section, we initially outline the three prerequisites for watermarks in NLG. Subsequently, we provide a synopsis of the two primary classifications of white-box watermarking strategies: post-hoc watermarking and inference-time watermarking.

\subsection{Watermarking Requirements}
Building on prior research in traditional digital watermarking, we put forth three crucial requirements for NLG watermarking  (1) Effectiveness: The watermark must be effectively embedded into the generated texts and verifiable while preserving the quality of the generated texts. (2) Secrecy: To achieve stealthiness, the watermark should be designed without introducing conspicuous alterations that could be readily detected by automated classifiers. Ideally, it should be indistinguishable from non-watermarked texts. (3) Robustness: The watermark ought to be resilient and difficult to remove through common modifications such as synonym substitution. To eliminate the watermark, attackers would need to implement significant modifications that render the texts unusable. These three requirements form the bedrock for NLG watermarking and guarantee the traceability of LLM-generated texts.

\subsection{Post-hoc Watermarking}
Given an LLM-generated text, post-hoc watermarks will embed a hidden message or identifier into the text. Verification of the watermark can be performed by recovering the hidden message from the suspicious text. There are two main categories of post-hoc watermarking methods: rule-based and neural-based approaches.

\subsubsection{Rule-based Approaches}
Initially, nature language researchers adapted techniques from multimedia watermarking, which were non-linguistic in nature and relied heavily on character changes. For example, the line-shift watermark method involves moving a line of text upward or downward (or left or right) based on the binary signal (watermark) to be inserted  \cite{brassil1995electronic}. However, these "printed text" watermarking approaches had limited applicability and were not robust against text reformatting. Later research shifted towards using the syntactic structure for watermarking. A study by Atallah et al. \cite{atallah2001natural} embedded watermarks in parsed syntactic tree structures, preserving the meaning of the original texts and rendering watermarks indecipherable to those without knowledge of the modified tree structure. Additionally, syntactic tree structures are difficult to remove through editing and remain effective when the text is translated into other languages. Further improvements were made in a series of works, which proposed variants of the method that embedded watermarks based on synonym tables instead of just parse trees \cite{jalil2009review}. Along with syntactic structure, researchers have also leveraged the semantic structure of text to embed watermarks. This includes exploiting features such as verbs, nouns, prepositions, spelling, acronyms, grammar rules, etc. For instance, a synonym substitution approach was proposed in which watermarks are embedded by replacing certain words with their synonyms without altering the context of the text \cite{topkara2006hiding}. Generally, rule-based methods utilize fixed rule-based substitutions, which may systematically change the text statistics, compromising the watermark's secrecy and allowing adversaries to detect and remove the watermark.

\subsubsection{Neural-based Approaches} 
In contrast to the rule-based methods that demand significant engineering efforts for design, neural-based approaches envision the information-hiding process as an end-to-end learning process. Typically, these approaches involve three components: a watermark encoder network, a watermark decoder network, and a discriminator network \cite{abdelnabi2021adversarial}. Given a target text and a secret message (e.g., random binary bits), the watermark encoder network generates a modified text that incorporates the secret message. The watermark decoder network then endeavors to retrieve the secret message from the modified text. One challenge is that the watermark encoder network may significantly alter the language statistics. To address this problem, the framework employs an adversarial training strategy and includes the discriminator network. The discriminator network takes the target texts and watermarked texts as input and aims to differentiate between them, while the watermark encoder network aims to make them indistinguishable. The training process continues until the three components achieve a satisfactory level of performance. For watermarking LLM-generated text, developers can use the watermark encoder network to embed a pre-set secret message into LLMs' outputs, and the watermark decoder network to verify suspicious texts. Although neural-based approaches eliminate the need for manual rule design, their inherent lack of interpretability raises concerns regarding their truthfulness and the absence of mathematical guarantees for the watermark's effectiveness, secrecy, and robustness.

\begin{figure}[t]
\centering
\includegraphics[width=1.0\columnwidth]{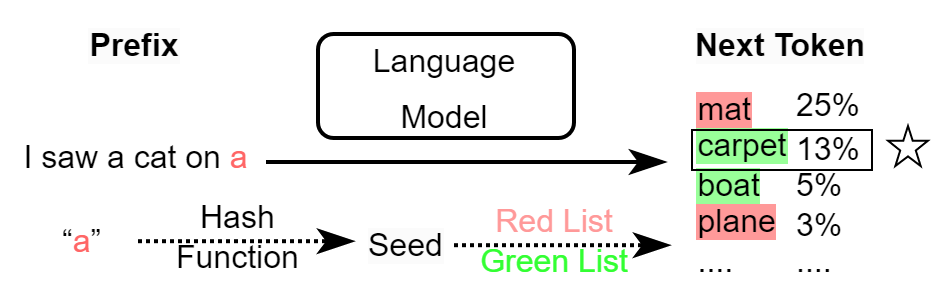}
 \caption{Illustration of inference time watermark. A random seed is generated by hashing the previously predicted token "a", splitting the whole vocabulary into "green list" and "red list". The next token "carpet" is chosen from the green list.}
 \label{fig: inference time watermark}
\end{figure}

\subsection{Inference Time Watermark}
Inference-time watermarking targets the LLM decoding process, as opposed to post-hoc watermarks, which are applied after text generation. The language model produces a probability distribution for the subsequent word in a sequence based on preceding words. A decoding strategy, which is an algorithm that chooses words from this distribution to create a sequence, offers an opportunity to embed the watermark by modifying the word selection process.

A representative example of this method can be found in research conducted by Kirchenbauer et al. \cite{kirchenbauer2023watermark}. During the next token generation, a hash code is generated based on the previously generated token, which is then used to seed a random number generator. This seed randomly divides the whole vocabulary into a "green list" and a "red list" of equal size. The next token is subsequently generated from the green list. In this way, the watermark is embedded into every generated word, as depicted in Figure. \ref{fig: inference time watermark}. To detect the watermark, a third party with knowledge of the hash function and random number generator can reproduce the red list for each token and count the number of violations of the red list rule, thus verifying the authenticity of the text. The probability that a natural source produces $N$ tokens without violating the red list rule is only $1/2^N$, which is vanishingly small even for text fragments with a few dozen words. To remove the watermark, adversaries need to modify at least half of the document's tokens. However, one concern with these inference-time watermarks is that the controlled sampling process may significantly impact the quality of the generated text. One solution is to relax the watermarking constraints, e.g., increasing the green list vocabulary size, and aim for a balance between watermarking and text quality.

\section{Benchmarking Datasets}
This section introduces several noteworthy benchmarking datasets for detecting LLM-generated texts. One prominent example is the work by Guo et al. \cite{guo2023close}, where the authors constructed the dataset Human ChatGPT Comparison Corpus (HC3), specifically designed to distinguish text generated by ChatGPT in question-answering tasks in both English and Chinese languages. The dataset comprises 37,175 questions across diverse domains, such as open-domain, computer science, finance, medicine, law, and psychology. These questions and corresponding human answers were sourced from publicly available question-answering datasets and wiki text. The researchers obtained responses generated by ChatGPT by presenting the same questions to the model and collecting its outputs. The evaluation results demonstrated that a RoBERTa-based detector achieved the highest performance, with F1 scores of 99.79\% for paragraph-level detection and 98.43\% for sentence-level detection of English text. Table \ref{tab:benchmark} presents more representative public benchmarking datasets for detecting different LLMs across various domains.

Despite numerous LLM-generated text detection studies, there is no comprehensive benchmarking dataset for performance comparison. This gap arises from the divergence in detection targets across different studies, focusing on distinct LLMs in various domains like news, question-answering, coding, and storytelling. Moreover, the rapid evolution of LLMs must be acknowledged. These models are being developed and released at an accelerating pace, with new models emerging monthly. As a result, it is increasingly challenging for researchers to keep up with these advancements and create datasets that accurately reflect each new model. Thus, the ongoing challenge in the field lies in establishing comprehensive and adaptable benchmarking datasets to accommodate the rapid influx of new LLMs.

\begin{table}[h]
    \centering
    \begin{tabular}{l|c|c}
    \toprule
        Dataset & LLMs & Domain \\ \hline
        HC3 \cite{guo2023close} & ChatGPT & Question Answering \\ \hline
        Neural Fake News \cite{zellers2019defending} & Grover & News \\ \hline
        TweepFake \cite{fagni2021tweepfake} & GPT2 & Tweets \\ \hline
        GPT2-Output \cite{openaiGPT2detection} & GPT2 & WebText  \\ \hline
        TURINGBENCH \cite{uchendu2021turingbench} & GPT1,2,3 & News \\ \hline
    \end{tabular}
    \caption{Representative Benchmarking Datasets}
    \label{tab:benchmark}
\end{table}
\section{Adaptive Attacks for Detection Systems}
\begin{figure}[t]
\centering
\includegraphics[width=1.0\columnwidth]{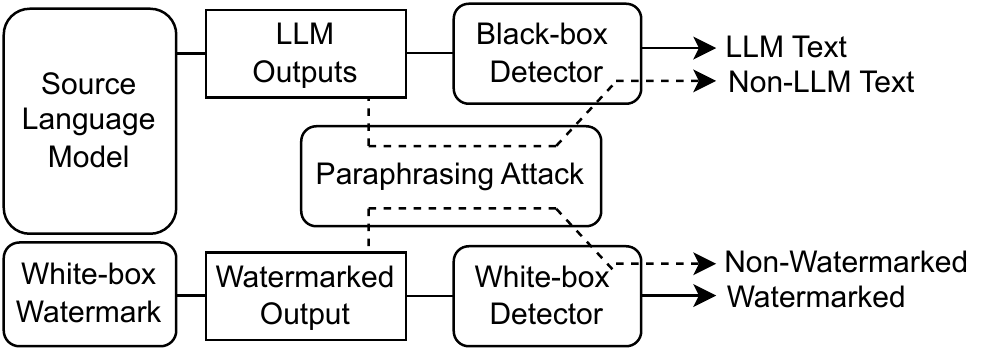}
 \caption{Illustration of paraphrasing attack.}
 \label{fig: paraphrasing attack}
\end{figure}

The previous section delineated both white-box and black-box detection methodologies.  This section pivots towards adversarial perspectives, delving into potential adaptive attack strategies capable of breaching existing detection approaches. A representative work is from Sadasivan et al. \cite{sadasivan2023can}. The authors empirically demonstrated that a paraphrasing attack could break a wide array of detectors, including both white-box and black-box approaches. This attack is premised on a straightforward yet potent assumption: given a sentence $s$, we denote $P(s)$ as the set of all paraphrased sentences that have similar meanings to $s$. Furthermore, let $L(s)$ represent the set of sentences that the LLM could generate with a similar meaning to $s$. We assume that a detector can only identify sentences within $L(s)$. If the size of $L(s)$ is much smaller than $P(s)$, i.e., $|L(s)| \ll |P(s)|$, attackers can randomly sample a sentence from $P(s)$ to evade the detector with a high probability. Based on this assumption, the author utilizes a different language model to paraphrase the output of the LLM, thereby simulating the sampling process from $P(s)$, as depicted in Figure \ref{fig: paraphrasing attack}. The empirical result shows that the paraphrasing attack is effective for the inference time watermark attack \cite{kirchenbauer2023watermark}. Utilizing the PEGASUS-based paraphrasing, the author succeeded in reducing the green list tokens from 58\% to 44\%. As a result, the detector accuracy drops from 97\% to 80\%. The attack also adversely affected black-box detection methods, causing the true positive rate of OpenAI's RoBERTa-Large-Detector to decline from 100\% to roughly 80\% with a practical false positive rate of 1\%. Future research will inevitably encounter a proliferation of attack strategies designed to dupe detection systems. Developing robust detection systems capable of withstanding such potential attacks poses a formidable challenge to researchers.

\section{Authors' Concerns} 

\subsection{Limitations of Black-box Detection} 

\textbf{Bias in Collected Datasets.} Data collection plays a vital role in the development of black-box detectors, as these systems rely on the data they are trained on to learn how to identify detection signals. However, it is important to note that the data collection process can introduce biases that can negatively impact the performance and generalization of the detector \cite{pagnoni2022threat}. These biases can take several forms. For example, many existing studies tend to focus on only one or a few specific tasks, such as question-answering or news generation, which can lead to an imbalanced distribution of topics in the data and limit the detector's ability to generalize. Additionally, human artifacts can easily be introduced during data acquisition, as seen in the study conducted by Guo et al. \cite{guo2023close}, where the lack of style instruction in collecting LLM-generated answers led to ChatGPT producing answers with a neutral sentiment. These spurious correlations can be captured and even amplified by the detector, leading to poor generalization performance in real-world applications. 

\noindent\textbf{Confidence Calibration.} In the development of real-world detection systems, it's crucial not only to have accurate classifications but also to provide an indication of the likelihood of being incorrect. For instance, a text with a 98\% probability of being generated by an LLM should be considered more likely to be machine-generated than one with a 90\% probability. In other words, the predicted class probabilities should reflect its ground truth correctness likelihood. Accurate confidence scores are of paramount importance for assessing system trustworthiness, as they offer valuable information for users to establish trust in the system, particularly for neural networks whose decisions can be challenging to interpret. Although neural networks exhibit greater accuracy than traditional classification models, research on confidence score accuracy in LLM-generated text detection topics remains scarce. Therefore, it is essential to calibrate the confidence scores for black-box detection classifiers, which frequently employ neural-based models.

In our opinion, while black-box detection works at present due to detectable signals left by language models in generated text, it will gradually become less viable as language model capabilities advance and ultimately become infeasible. In light of the rapid improvement in LLM-generated text quality, the future of reliable detection tools lies in white-box watermarking detection approaches.

\subsection{Limitations of White-box Detection}
The limitations of the white-box detection method primarily stem from two perspectives. Firstly, there exists a trade-off between the effectiveness of the watermark and the quality of the text, which applies to both post-hoc watermarks \cite{jalil2009review, abdelnabi2021adversarial} and inference time watermarks \cite{sadasivan2023can}. Achieving a more reliable watermark often requires significant modifications to the original text, potentially compromising its quality. The challenge lies in optimizing the trade-off between watermark effectiveness and text quality to identify an optimal balance. Moreover, as users accumulate a growing volume of watermarked text, there is a potential for adversaries to detect and reverse engineer the watermark. Sadasivan et al. \cite{sadasivan2023can} emphasized that when attackers query the LLMs, they can gain knowledge about the green list words proposed in the inference time watermark \cite{kirchenbauer2023watermark}. This enables them to generate text that fools the detection system. To address this issue, it is essential to quantify the detectability of the watermark and explore its robustness under different query budgets. These avenues present promising directions for future research.

\subsection{Lacking Comprehensive Evaluation Metrics}

Existing studies often rely on metrics such as  AUC or accuracy for evaluating detection performance. However, these metrics only consider an average case and are not enough for security analysis. Consider comparing two detectors: Detector A perfectly identify of 1\% of the LLM-generated texts but succeeds with a random 50\% chance on the rest. Detector B succeeds with 50.5\% on all data. On average, two detectors have the same detection accuracy or AUC. However, detector A demonstrates exceptional potency, while detector B is practically ineffective. In order to know if the detector can reliably identify the LLM-generated text, researchers need to consider the low false-positive rate regime (FPR) and report a detector's True-Positive Rate (TPR) at a low false-positive rate. This objective of designing methods around low false-positive regimes is widely used in the computer security domain \cite{kantchelian2015better}. This is especially crucial for populations who  produce unusual text, such as non-native speakers. Such populations might be especially at risk for false-positive, which could lead to serious consequences if these detectors are used in our education systems.

\subsection{Threats from Open-Source LLMs} 

Current detection methods are based on the assumption that the LLM is controlled by the developers and offered as a service to end-users, this one-to-many relationship is conducive to detection purposes. However, challenges arise when developers open-source their models or when hackers steal them. For instance, Meta's latest LLM, LLaMA, was initially accessible by request. But just a week after accepting access requests, it was leaked online via a 4chan torrent, raising concerns about the potential surge in personalized spam and phishing attempts \cite{meta2023leak}. Once the end user gets full access to the LLM, the ability to modify the LLMs' behavior hinders black-box detection from identifying generalized language signals. Embedding watermarks in LLM outputs is one potential solution, as developers can integrate concealed watermarks into LLM outputs before making the models open-source. However, it can still be defeated as users have full access to the model and can fine-tune it or change sampling strategies to erase existing watermarks. 
Furthermore, it is challenging to impose the requirement of embedding a traceable watermark on all open-source LLMs. 

With a growing number of companies and developers engaging in open-source large language model projects, a future possibility emerges wherein individuals can own and customize their own language models. In this scenario, the detection of open-source LLMs becomes increasingly complex and challenging. As a result, the threat emanating from open-source LLMs demands careful consideration, given its potentially wide-ranging implications for the security and integrity of various sectors. 

\section{Conclusion}
The detection of LLM-generated texts is an expanding and dynamic field, with numerous newly developed techniques emerging continuously. This survey provides a precise categorization and in-depth examination of existing approaches to help the research community comprehend the strengths and limitations of each method. Despite the rapid advancements in LLM-generated text detection, significant challenges still need to be addressed. Further progress in this field will require developing innovative solutions to overcome these challenges.
\newpage
\bibliographystyle{ACM-Reference-Format}
\bibliography{ref}
\end{document}